\documentclass[10pt,twocolumn,letterpaper]{article}

\usepackage[pagenumbers]{iccv} 
\usepackage{algorithm}
\usepackage{algorithmic}
\usepackage{caption}
\usepackage{setspace}
\definecolor{iccvblue}{rgb}{0.21,0.49,0.74}
\usepackage[pagebackref,breaklinks,colorlinks,allcolors=iccvblue]{hyperref}
\usepackage{multirow}
\def\paperID{15148} 
\def\confName{ICCV}
\def\confYear{2025}

\title{MsaMIL-Net: An End-to-End Multi-Scale Aware Multiple Instance Learning Network for Efficient Whole Slide Image Classification}

\author{Jiangping Wen$^{1}$ \quad Jinyu Wen$^{*,1}$ \quad Meie Fang$^{1}$\\
$^{*}$Corresponding Author\\
$^{1}$Guangzhou University\\
Guangzhou 511400, China\\
{\tt\small 1462887221@qq.com, wjy1361120721@163.com, fme@gzhu.edu.cn}
}

\begin{document}
\maketitle
\begin{abstract}
Bag-based Multi-Instance Learning (MIL) methods have become the mainstream approach for Whole Slide Image (WSI) classification. However, most existing methods adopt a segmented training strategy, which first extracts features using a pre-trained feature extractor, and then uses MIL for feature aggregation. This segmented training approach leads to insufficient collaborative optimization between the feature extraction network and MIL network, preventing end-to-end joint optimization and thus limiting the overall performance of the model. Moreover, existing methods typically extract features from all patches of fixed size, ignoring the multi-scale observation characteristics of pathologists, which not only results in serious computational resource waste when the proportion of tumor regions is extremely low (such as in the Camelyon16 dataset), but may also cause the model to converge to suboptimal solutions.

To address these issues, this paper proposes an end-to-end multi-scale WSI classification framework that combines multi-scale feature extraction with multi-instance learning. Specifically, we design the following modules: \textbf{(1)} Semantic Feature Filtering Module, which uses UNet++ to segment lesion regions at low magnification, reducing interference from non-lesion regions and significantly decreasing computational load; \textbf{(2)} Multi-scale Feature Extraction Module, which extracts multi-scale features using native patches at 20x, 10x, and 5x magnifications to capture pathological information at different levels; \textbf{(3)} Multi-scale Fusion MIL Module for global modeling and efficient fusion of multi-scale features. Additionally, we propose an end-to-end training strategy that simultaneously optimizes the feature extractor and MIL network through techniques such as gradient accumulation, ensuring maximum compatibility between the two components.

Experiments were conducted on three cross-center datasets (DigestPath2019, BCNB, and UBC-OCEAN). Extensive experimental results show that our method outperforms existing state-of-the-art approaches in terms of both accuracy (ACC) and AUC metrics.
\end{abstract}    
\section{Introduction}
\label{sec:intro}

Pathological analysis is the "gold standard" for cancer diagnosis, and Whole Slide Images (WSIs) as the core tool for pathological analysis present significant challenges to traditional deep learning methods due to their high resolution and enormous size \cite{qu2022towards,srinidhi2021deep,mahmood2019deep,lu2021ai,lu2022federated}. Since WSI images cannot be processed completely at once, bag-based Multiple Instance Learning (MIL) methods have become the mainstream solution for WSI classification.
 Since WSI images cannot be processed completely at once, bag-based Multiple Instance Learning (MIL) methods have become the mainstream solution for WSI classification. Under the MIL framework, a WSI is viewed as a "bag" containing multiple "instances" (i.e., patches), with only bag-level labels available, making MIL a weakly supervised learning method. Currently, mainstream MIL methods typically first extract features using pre-trained feature extractors and then aggregate features through MIL networks \cite{ilse2018attention,shao2021transmil,yang2024mambamil,fang2024mammil,qu2022dgmil,qu2024rethinking,qu2022bi,hashimoto2020multi}.
 However, existing open-source feature extractors are usually pre-trained only at fixed magnifications (e.g., 20x), and there are almost no open-source multi-scale feature extractors, so these methods cannot leverage multi-scale features. Research shows that multi-scale methods \cite{hashimoto2020multi,deng2022cross,bontempo2023graph} can significantly
improve WSI image classification performance, further verifying the importance of multi-scale features. Moreover, existing pre-training methods (such as minimizing reconstruction loss in pixel space and contrastive learning), while performing well on general tasks, still require fine-tuning on corresponding cancer datasets to achieve optimal performance in specific WSI image classification tasks. To improve WSI image classification performance and fully utilize multi-scale features, this paper proposes an end-to-end training method and designs a multi-scale feature extractor network. Unlike existing MIL methods (such as TransMIL \cite{shao2021transmil}, MambaMIL \cite{yang2024mambamil}, and other representative approaches \cite{ilse2018attention,qu2022dgmil,qu2024rethinking,qu2022bi})
), which mainly classify features extracted at fixed magnifications and cannot effectively aggregate multi-scale information, this paper further designs a new MIL network that achieves efficient fusion of multi-scale features by introducing a multi-scale feature fusion mechanism, thereby enhancing classification performance.

WSI images contain a large amount of redundant tissue,with some datasets \cite{bejnordi2017diagnostic} containing less than 10\% 
 tumor tissue. WSI images are enormous, and extracting features and performing MIL feature aggregation for all patches leads to tremendous resource waste. Some works [9,10] indicate that simple segmentation processing of lesion areas can be achieved at low magnification. Traditional methods involve fixed 20x magnification cropping of 256px patches for feature extraction followed by multiple instance learning, preventing models from obtaining higher-scale features such as complete lymphatic vessel distribution and other macro-scale structures, and only allowing integration of cellular or subcellular level features for classification. Pathologists' analytical method involves preliminary judgment of lesion areas under low magnification, then observing tissue structures (such as lymphatic vessels) under medium magnification, and finally analyzing nuclear details (such as nucleus-to-cytoplasm ratio) under high magnification, ultimately determining categories through multi-scale combination. This work incorporates similar thinking, first pre-segmenting WSI images for cancer at less than 1x magnification at the thumbnail level, simulating pathologists' preliminary screening of disease areas under low magnification, then simulating pathologists' multi-scale comprehensive judgment of different tissues under medium and high magnification, using features of different scales for multiple instance learning to ultimately complete the classification of WSI images.

\section{related work}
\label{sec:formatting}

\subsection{Self-supervised Feature Extractor Training Methods in MIL}

Due to the enormous size of WSI images, it is challenging to use the entire WSI for end-to-end training at once, and there is also a lack of pixel-level annotations. Therefore, the current mainstream approach is to pre-train feature extractors first, using self-supervised methods to train feature extractors such as DINO \cite{oquab2023dinov2}, PHIKON \cite{filiot2024phikon}, CTRANPATH \cite{wang2022transformer}, etc. Then, the trained feature
extractor is used to extract features from each cropped patch, these extracted features are saved, and subsequently, an MIL network is trained for classification. However, this "disconnected" segmented training approach cannot achieve simultaneous optimization of the feature extraction network and the MIL network. The optimization objectives of these two networks are separated, making it impossible to achieve global optimization.

\subsection{Current Multi-scale MIL Methods}

In terms of multi-scale feature extraction and fusion, existing methods also have significant limitations. For example,CSMIL \cite{deng2022cross} separately trains feature extractors for three scales, but only uses convolutional gated attention mechanisms for multi-scale fusion, failing to achieve deep global modeling. The method of Zhang et al. \cite{zhang2022self} merely averages prediction probabilities across different scales without establishing multi-scale correlations at the feature level. While MILBooster \cite{qu2023boosting} increases the proportion of positive instances through a "bag filter," its selection method based on feature clustering and Mahalanobis distance is only applicable to binary classification datasets and difficult to extend to multi-class tasks. Additionally, this method only uses dual-scale fusion, resulting in poor feature consistency when there are large scale jumps. Furthermore, HIPT [8] implements multi-scale modeling through a hierarchical self-supervised learning framework, but its high-level features depend on rigid concatenation of low-magnification features, making it difficult to effectively capture native macroscopic pathological patterns (such as the continuous morphological features of lymphatic vessel infiltration).

\subsubsection{Limitations of Existing Methods and Our Solutions}

The limitations of existing methods can be summarized as follows: \textbf{(1)} Multi-scale learning requires using self-supervised methods to train separate feature extraction models for different scales, resulting in high training costs. Additionally, the segmented training strategy decouples the optimization objectives of the feature extraction network and the MIL network, preventing collaborative optimization between the two and leading to insufficient adaptation of feature representation to classification tasks, ultimately causing models to converge to suboptimal solutions. \textbf{(2)} Existing methods require feature extraction for all patches, and even MILBooster needs to filter on already extracted features, resulting in significant waste of computational resources. \textbf{(3)} Most methods only use dual-stream, bi-scale approaches, with large scale jumps causing scale gap issues. \textbf{(4)} Some methods fail to properly model relationships between multi-scale features. \textbf{(5)} Some methods do not extract multi-scale image features in the native pixel domain but rely on concatenating extracted feature vectors from lower scales, making it difficult to capture native macroscopic pathological features. \textbf{(6)} The vast majority of methods do not filter out interfering tissue regions, potentially affecting final classification. MILBooster's "bag filter" method can only be used with binary classification datasets that have positive and negative bag distinctions, posing significant limitations.

\subsection{Method Design and Innovations}
To address problems 2 and 6, we chose to use UNet++ \cite{zhou2018unet++} at the thumbnail level to pre-segment WSIs to reduce processing time and interference from non-lesion regions. This preliminary filtering of potentially interfering non-lesion regions allows us to extract features only from lesion areas, providing better filtering of interfering instances compared to MILBooster's "bag filter" approach, and it can be applied to multi-class datasets. For problems 3, 4, and 5, we use native patch images at three scales—20x, 10x, and 5x—to avoid scale gap issues and to extract native microscopic and macroscopic pathological features. Through our multi-scale aggregation module, we achieve global modeling of multi-scale features. For problem 1, we propose an end-to-end training strategy that synchronously optimizes the feature extractor and MIL network through techniques such as gradient accumulation, ensuring collaborative updates of both components under the guidance of a global loss function. This enables the feature extractor to dynamically adapt to the aggregation requirements of the MIL network, thereby enhancing overall model performance.

\section{method}
\begin{algorithm}
  \caption{WSI Image Classification Inference Pipeline}
  \label{alg:wsi_classification}
  \begin{algorithmic}
    \STATE \textbf{Input:} Whole Slide Image (WSI) $I \in \mathbb{R}^{H \times W \times 3}$
    \STATE \textbf{Output:} Predicted classification result $y_{\text{pred}}$ for WSI
    \STATE \textbf{1. Semantic Feature Filtering Module (SFFM)}
    \STATE \quad 1.1 Resize $I$ to a low-resolution image $I_{\text{low}}$ of size $1024 \times 1024$, with scaling factor $s = H/1024$
    \STATE \quad 1.2 Use a segmentation network $f_{\text{seg}}$ to generate a mask $M \in \{0,1\}^{1024 \times 1024 \times 3}$ from $I_{\text{low}}$
    \STATE \quad 1.3 Extract multi-scale patches (20x, 10x, 5x) from the original WSI image $I$
    \STATE \quad 1.4 Retain patches where the proportion of tumor area (red channel in mask $M$) exceeds 0.7, forming the final patch set $P$
    \STATE \textbf{2. Multi-Scale Feature Extraction Module (MSFEM)}
    \STATE \quad 2.1 Resize each patch $P_k$ in $P$ to $512 \times 512$, resulting in $P'_k = \text{Resize}(P_k, 512 \times 512)$
    \STATE \quad 2.2 Use a feature extractor $f_x$ to extract a $1 \times 512$ feature vector $C$ from each $P'_k$
    \STATE \textbf{3. Instance-Aware Attention Module (IAAM)}
    \STATE \quad 3.1 Explicitly inject relative position and scale information into the multi-scale patch features $T \in \mathbb{R}^{1 \times N \times 512}$, obtaining $T'$
    \STATE \quad 3.2 Process $T'$ using IAAM (including low-rank latent attention and dynamic multi-instance query mechanisms) for multi-scale fusion, obtaining the bag-level feature $F_{\text{bag}}$
    \STATE \quad 3.3 Feed $F_{\text{bag}}$ into the classification head to obtain the predicted classification probability $y_{\text{pred}}$ for the WSI
    \STATE \textbf{Return:} $y_{\text{pred}}$
  \end{algorithmic}
\end{algorithm}

\begin{figure*}[htbp]
      \centering
      \includegraphics[width=0.8\linewidth]{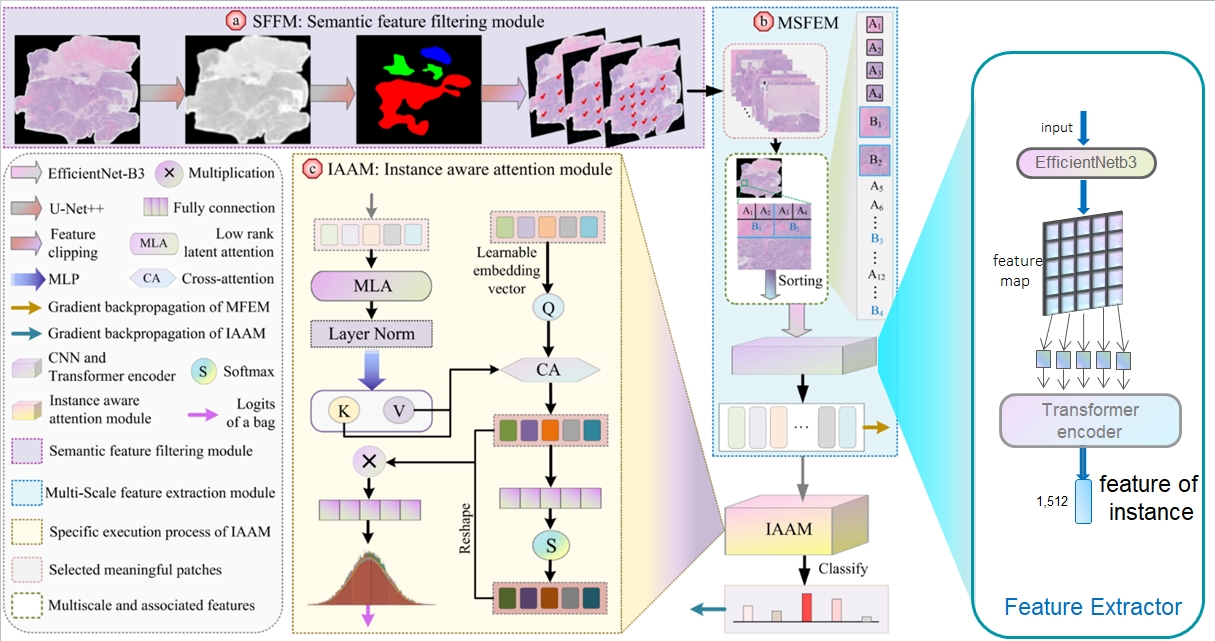}
      \caption{This paper introduces an end-to-end Multi-scale aware multiple instance learning network framework (MsaMIL-Net), which primarily comprises three modules: the semantic feature filtering module (SFFM), the multi-scale feature extraction module (MSFEM), and the instance-aware attention module (IAAM). Among them, SFFM can effectively reduce the interference of non-lesion areas in WSI classification tasks while enhancing inference speed and efficiency. MSFEM enables cross-scale semantic alignment, ranging from microscopic cellular morphology to macroscopic tissue architecture. IAAM enhances the model's capability to capture instance-level details, facilitating deep interaction across multi-scale information.}\label{framework}
    \end{figure*}

This paper proposes the first end-to-end differentiable multi-scale aware multiple instance learning network framework (MsaMIL-Net), which primarily consists of three modules: semantic feature filtering module (SFFM), multi-scale feature extraction module (MSFEM), and multi-scale feature fusion module (MSFM). As shown in Fig.~\ref{framework} and Algorithm~\ref{alg:wsi_classification}, these modules form a hierarchical feature refinement pipeline, enabling progressive reasoning from coarse-grained localization (lesion ROI extraction) to fine-grained discrimination (malignant instance filtering). Additionally, the entire model training process is end-to-end, primarily manifested in the gradient backpropagation algorithm between the feature extractor and the multiple instance learning classifier.

This paper proposes the first end-to-end differentiable multi-scale aware multiple instance learning network framework (MsaMIL-Net), which primarily consists of three modules: semantic feature filtering module (SFFM), multi-scale feature extraction module (MSFEM), and multi-scale feature fusion module (MSFM). As shown in Fig.~\ref{framework} and \cref{alg:wsi_classification}, these modules form a hierarchical feature refinement pipeline, enabling progressive reasoning from coarse-grained localization (lesion ROI extraction) to fine-grained discrimination (malignant instance filtering). Additionally, the entire model training process is end-to-end, primarily manifested in the gradient backpropagation algorithm between the feature extractor and the multiple instance learning classifier.

\subsection{Semantic Feature Filtering Module}
To effectively reduce the interference of non-lesion regions in WSI classification tasks while improving inference speed and efficiency, we propose the Semantic Feature Filtering Module (SFFM), as shown in Fig.~\ref{framework}. The workflow of this module is as follows:

First, the original WSI image $I$ is scaled to a resolution of $1024 \times 1024$, with scaling factors $s_1 = W/1024$ and $s_2 = H/1024$ for width and height, respectively. Next, a segmentation network based on UNet++ generates a mask $M$, where positions with a value of 1 in the red channel indicate lesion regions. On the mask image $M$, we perform cropping with fixed strides: horizontal stride $d_k/s_1$ and vertical stride $d_k/s_2$. For each cropped region, if the area proportion of the red channel exceeds 0.7, we map the coordinates back to the original image $I$, obtain the coordinates $(x,y)$ of the patch in the original image, and crop the patch according to the following formula:
\begin{equation}
P_k = I(x-d_k/2:x+d_k/2, y-d_k/2:y+d_k/2)
\end{equation}
where $d_k \in \{512\text{px}, 1024\text{px}, 2048\text{px}\}$ represents the size of the cropped patch. We save the cropped $P_k$, ultimately obtaining a collection $P$ of multi-scale lesion region patches of different sizes.

Through the Semantic Feature Filtering Module, we significantly reduce the number of instances that need to be processed in multiple instance learning, improve image processing speed, reduce computational cost, and minimize irrelevant interfering instances in the MIL network. The segmentation dataset is derived from lesion region masks provided by public datasets such as UBC-OCEAN. The segmentation dataset is derived from lesion region masks provided by public datasets such as UBC-OCEAN.The segmentation dataset comes from lesion region masks provided by public datasets such as UBC-OCEAN.
\subsection{Multi-scale Feature Extraction Module}
To overcome the challenges associated with independently training multi-scale feature extractors, we propose an innovative Multi-Scale Feature Extraction Module (MSFEM), as shown in Fig.~\ref{framework}(b). Traditional methods (such as CSMIL) pre-train feature extractors independently for different scales, resulting in heterogeneous feature distributions across scales, making cross-scale consistency modeling difficult and training costs expensive. In contrast, MSFEM achieves implicit cross-scale consistent representation learning through shared network parameters and end-to-end joint optimization. For example, high-magnification features automatically enhance complementarity with low-magnification structural features, effectively avoiding the scale gap issue.

First, each patch $P_k$ from the multi-scale patch collection $P$ retained in the previous stage is uniformly adjusted to a size of $512 \times 512$, representing 20x, 10x, and 5x magnification levels respectively, to capture lesion features at different hierarchical levels. Subsequently, the feature extractor $f_x$ processes each adjusted patch to obtain a $1 \times 512$ feature vector. Specifically, the adjusted patches are input into EfficientNetB3 to acquire the final layer feature map $F \in \mathbb{R}^{16 \times 16 \times 1536}$, expressed as $F = \text{EfficientNetB3}(P_k)$. The feature map $F$ is then reshaped into a sequence $S \in \mathbb{R}^{256 \times 1536}$, where each pixel point represents a token. Position encoding $E$ is added to sequence $S$ before being input into the Transformer Encoder, which outputs the class token $C \in \mathbb{R}^{1 \times 512}$ from the last layer, expressed as $C = \text{Transformer-Encoder}(S+E)$, where $C$ represents the feature of a patch.

Through this process, we extract $1 \times 512$ feature vectors from multi-scale patches. This design efficiently extracts spatial structural features within patches through CNN's convolutional structure, while the subsequent Transformer models global dependencies within patches through self-attention mechanisms, providing richer feature inputs for the subsequent multi-scale fusion MIL.
\subsection{IAAM: Instance Aware Attention Module}
Our IAAM is a multi-scale MIL method, subsequently referred to as multi-scale MIL, which is divided into Multi-scale Hierarchical Encoder (MHE) and Dynamic Multi-instance Query (DMQ). Addressing issues (3), (4), and (5) mentioned in related work, the MHE endpoint inputs pathological image features acquired from three native scales, achieving smooth transition of cross-scale information. Through the coordinate-scale encoding module, it explicitly injects spatial position and scale information of patches, and combines with low-rank latent attention layers to model global dependencies. The DMQ endpoint dynamically queries multi-scale instance features from the MHE endpoint, and finally fuses features of multiple scales through a gated attention layer, achieving comprehensive interaction and global modeling of multi-scale information.
\subsubsection{Multi-scale Hierarchical Encoder (MHE)}
For a bag containing patches of multiple scales, let $N_1$, $N_2$, and $N_3$ represent the number of patches at 20x, 10x, and 5x scales respectively, with a total patch count of $N = N_1 + N_2 + N_3$. After bag-level feature extraction, we obtain a feature collection $T \in \mathbb{R}^{1 \times N \times 512}$, composed of features $C \in \mathbb{R}^{1 \times 512}$ from each patch. For each patch, its top-left coordinate is $(x,y)$. Arranged in ascending order of $x$ and $y$ values, features from patches of different scales from the same or nearby regions will be adjacently arranged in the input sequence, as shown in module b of \cref{framework}. For features A, B, C of three different scales, they can be arranged as $\{A1, A2, A3, A4, B1, B2, \ldots, A12, \ldots, B4, \ldots\}$. This arrangement ensures the model can perceive spatial position relationships between patches, enhancing feature modeling capability. For the feature collection $T \in \mathbb{R}^{1 \times N \times 512}$, we explicitly inject each patch's relative position, scale information,and standard position encoding for transformers \cite{vaswani2017attention} $E \in \mathbb{R}^{1 \times N \times 512}$, obtaining the processed feature collection $T'$:
\begin{equation}
T' = T + FC([x, y, \text{scale}]) + E
\label{eq:feature_encoding}
\end{equation}
where scale is 0, 1, 2 representing the three scales 20x, 10x, 5x respectively, and the fully connected layer is $FC(3, 512)$.
Next, $T^{\prime}$ is input into the Multi-head Low-rank Attention (MLA) layer. In the MLA process, the Query and Key are first mapped to a low-rank latent space (with rank $r$, typically $r \ll d$):

\begin{equation}
Q_{low} = T^{\prime}W_Q^{low}, \quad K_{low} = T^{\prime}W_K^{low}
\end{equation}

where $W_Q^{low}, W_K^{low} \in \mathbb{R}^{512 \times r}$ and $W_V \in \mathbb{R}^{512 \times 512}$. The attention weights are then calculated in the low-rank latent space:

\begin{equation}
A_{low} = \text{Softmax}\left(\frac{Q_{low}K_{low}^T}{\sqrt{r}}\right) \in \mathbb{R}^{N \times N}
\end{equation}

These low-rank attention weights are used to weight the original values $V$, generating an intermediate output $T_{inter} = A_{low}V \in \mathbb{R}^{1 \times N \times 512}$. Layer Normalization is then applied to this intermediate output. Finally, the multi-scale features are further processed through an MLP (Multi-Layer Perceptron) to obtain the final output:

\begin{equation}
T^{\prime\prime\prime} = \text{MLP}(\text{LayerNorm}(A_{low}(T^{\prime}W_V)))
\end{equation}
\subsubsection{Dynamic Multi-instance Query (DMQ)}
In the MHE endpoint features $T'''$, there exists an imbalance in the number of instances between scales: the number of 20x patches is far greater than 5x, causing the model to be easily dominated by high-resolution patches during fusion, ignoring global semantic information at macro scales. To address this issue, we designed a dynamic query mechanism at the DMQ endpoint, using learnable embedding vectors $Z \in \mathbb{R}^{10 \times 512}$ as queries to interact with the multi-scale features $T'''$ output from the MHE endpoint through cross-attention, allowing the model to autonomously select key features from which scales (20x/10x/5x) and spatial positions. This mechanism dynamically balances the contribution of patches at different scales, mitigating the problem of high-resolution patch dominance, while allowing the model to autonomously select features from key scales and spatial positions based on task requirements, automatically filtering lesion areas strongly correlated with the classification task and suppressing interference from non-lesion areas. In technical implementation, the learnable embedding vector $Z \in \mathbb{R}^{10 \times 512}$ first generates a query matrix $Q_Z = ZW_Q$ through a linear layer, where $W_Q \in \mathbb{R}^{512 \times 512}$. Then, key matrix $K = T''' W_K$ and value matrix $V = T''' W_V$ are extracted from the MHE endpoint feature $T'''$. Finally, refined multi-scale features $Z' \in \mathbb{R}^{10 \times 512}$ are calculated through cross-attention, with the formula:
\begin{equation}
Z' = \text{Softmax}((Q_Z K^T) / \sqrt{d})V
\label{eq:cross_attention}
\end{equation}
The Gated Attention Layer, as the final layer of the DMQ module, assigns weights to features at different scales after obtaining the features $Z^{\prime}$, reflecting their importance in the classification task. First, for each scale's feature vector $Z_i^{\prime} \in \mathbb{R}^{1 \times 512}$ (where $i=1,2,\ldots,10$), gating weights are generated:
\begin{equation}
g_i = \sigma(W_g Z_i^{\prime} + b_g)
\end{equation}
where $W_g$ and $b_g$ are learnable parameters, and $\sigma$ is an activation function. Then, the feature vectors are weighted and summed using the gating weights to obtain the bag-level feature:
\begin{equation}
F_{bag} = \sum_{i=1}^{10} g_i \cdot Z_i^{\prime}
\end{equation}
Finally, $F_{bag} \in \mathbb{R}^{1 \times 512}$ serves as the bag-level feature and is input to the classification head to complete WSI classification.
\subsection{End-to-End Feature Extractor Training Process}
\begin{algorithm}
\caption{End-to-End Training Pipeline}
\label{alg:e2e_training}
\begin{algorithmic}[1]
\REQUIRE Training set of WSIs with bag-level labels
\ENSURE Optimized parameters for feature extractor and MIL network
\STATE \textbf{Input:} $P^{\text{all}} = \{P^{\text{all}}_1, \ldots, P^{\text{all}}_K\}$
\STATE \textbf{Batch Selection:} $P_B \subset P^{\text{all}}$, $|P_B| = 200$
\STATE \textbf{Feature Extraction:} $C_B = \{f(P_b) | P_b \in P_B\}$, where $f(P_b) = \text{extractor}(P_b; \theta_{\text{extractor}})$
\STATE \textbf{MIL Prediction:} $y_{\text{pred}} = \text{MIL}(C_B; \theta_{\text{MIL}})$
\STATE \textbf{Loss Calculation:} $L = \text{CrossEntropy}(y_{\text{pred}}, y_{\text{WSI}})$
\STATE \textbf{Gradient Update:}
\STATE \quad $\theta_{\text{extractor}} \leftarrow \theta_{\text{extractor}} - \eta\nabla_{\theta_{\text{extractor}}}L$
\STATE \quad $\theta_{\text{MIL}} \leftarrow \theta_{\text{MIL}} - \eta\nabla_{\theta_{\text{MIL}}}L$
\RETURN Optimized parameters $\theta_{\text{extractor}}$ and $\theta_{\text{MIL}}$
\end{algorithmic}
\end{algorithm}
Addressing issue (1), namely the inconsistent optimization objectives between feature extraction networks and MIL networks in segmented training, we propose an end-to-end training strategy. By jointly optimizing the feature extractor and MIL network, we ensure they update collaboratively under the guidance of a global loss function, thereby achieving maximum alignment between the feature space and the classification task.
For a WSI image $I \in \mathbb{R}^{H \times W \times 3}$, to enable the feature extractor to learn to extract features from all patches, we select all patches, i.e., multi-resolution patches cropped from all non-background areas (including lesions, stroma, and irrelevant tissues), denoted as set $P^{all} = \{P_{1,r}^{all}, P_{2,r}^{all}, \ldots, P_{N,r}^{all}\}$, where $r \in \{512, 1024, 2048\}$ represents patch resolution (px), and $N$ is the total number of patches (e.g., $N \approx 10000$). In end-to-end training, we do not save intermediate feature files but instead process dynamically: randomly select 200 patches from $P^{all}$ as a batch $P_B = \{P_1, P_2, \ldots, P_{200}\}$, resize each patch, extract features through the feature extractor to obtain a feature collection $C_B = \{C_1, C_2, \ldots, C_{200}\}$, where $C_b \in \mathbb{R}^{1 \times 512}$. Input $C_B$ into the multi-scale fusion MIL network to obtain the predicted classification probability of the WSI image $y_{pred} = \text{MIL}(C_B; \theta_{MIL})$, where $y_{pred} \in \mathbb{R}^C$ represents class probabilities, and $C$ is the number of classes. By calculating cross-entropy loss $L = \text{CrossEntropy}(y_{pred}, y_{WSI})$, backpropagate gradients and update parameters of the feature extractor and MIL network:
\begin{equation}
\begin{aligned}
\theta_{extractor} &\leftarrow \theta_{extractor} - \eta \nabla_{\theta_{extractor}} L, \\
\theta_{MIL} &\leftarrow \theta_{MIL} - \eta \nabla_{\theta_{MIL}} L
\end{aligned}
\label{eq:parameter_updates}
\end{equation}
where $\eta$ is the learning rate. To ensure gradients can propagate back to the feature extractor, the computational graph retains the feature extraction and MIL prediction processes for 200 patches, ensuring end-to-end differentiability. Compared to two-stage methods, end-to-end training dynamically optimizes the feature extractor and MIL network, with feature representation highly adapted to the classification task; two-stage methods require extracting and saving features of all patches first, then training the MIL network, decoupling feature representation from the classification task.

Through the end-to-end training framework, the feature extractor can efficiently extract multi-scale features, providing high-quality input for multiple instance learning. During training, we employ gradient checkpointing and gradient accumulation techniques, successfully achieving simultaneous training of the feature extractor and MIL network with a single batch instance count of $B = 200$, and efficiently completing training on two NVIDIA 3080 GPUs. Subsequently, to maximize classification accuracy, we further train the MIL network separately, using the trained multi-scale feature extractor to extract features from all lesion area patches in the WSI image, and save the features in .pt format. Afterward, based on features from an average of 3000 patches of different scales containing only lesion areas, we further train the multi-scale MIL network to achieve optimal classification performance. This additional MIL network training is conducted in all subsequent comparative and ablation experiments.

\section{Experiments and Results}

\subsection{Datasets and Implementation Details}
To validate the performance of our architecture, we conducted experiments on three public datasets: DigestPath2019 \cite{da2022digestpath}, BCNB, and UBC-OCEAN \cite{farahani2022deep,asadi2024machine}.
 DigestPath2019 is a dataset for colonoscopy pathology examination, with WSI images scanned at 20x magnification and containing pixel-level tumor region annotation masks, which we used for binary classification tasks (benign vs. malignant). The BCNB dataset focuses on predicting early breast cancer lymph node metastasis, containing partial tumor region annotations, which we used for classifying lymph node metastasis status. The UBC-OCEAN dataset contains more than five hundred ovarian cancer subtype images, some with tumor region annotations, which we used for a five-class classification task of ovarian cancer subtypes. Since our architecture includes a pre-segmentation network that can quickly screen lesion areas, we trained the segmentation network using the tumor region annotations provided by the three datasets mentioned above.
\subsection{Ablation Study on Lesion Area Segmentation}
   \begin{figure}[htbp]
      \centering
      \includegraphics[width=0.7\linewidth]{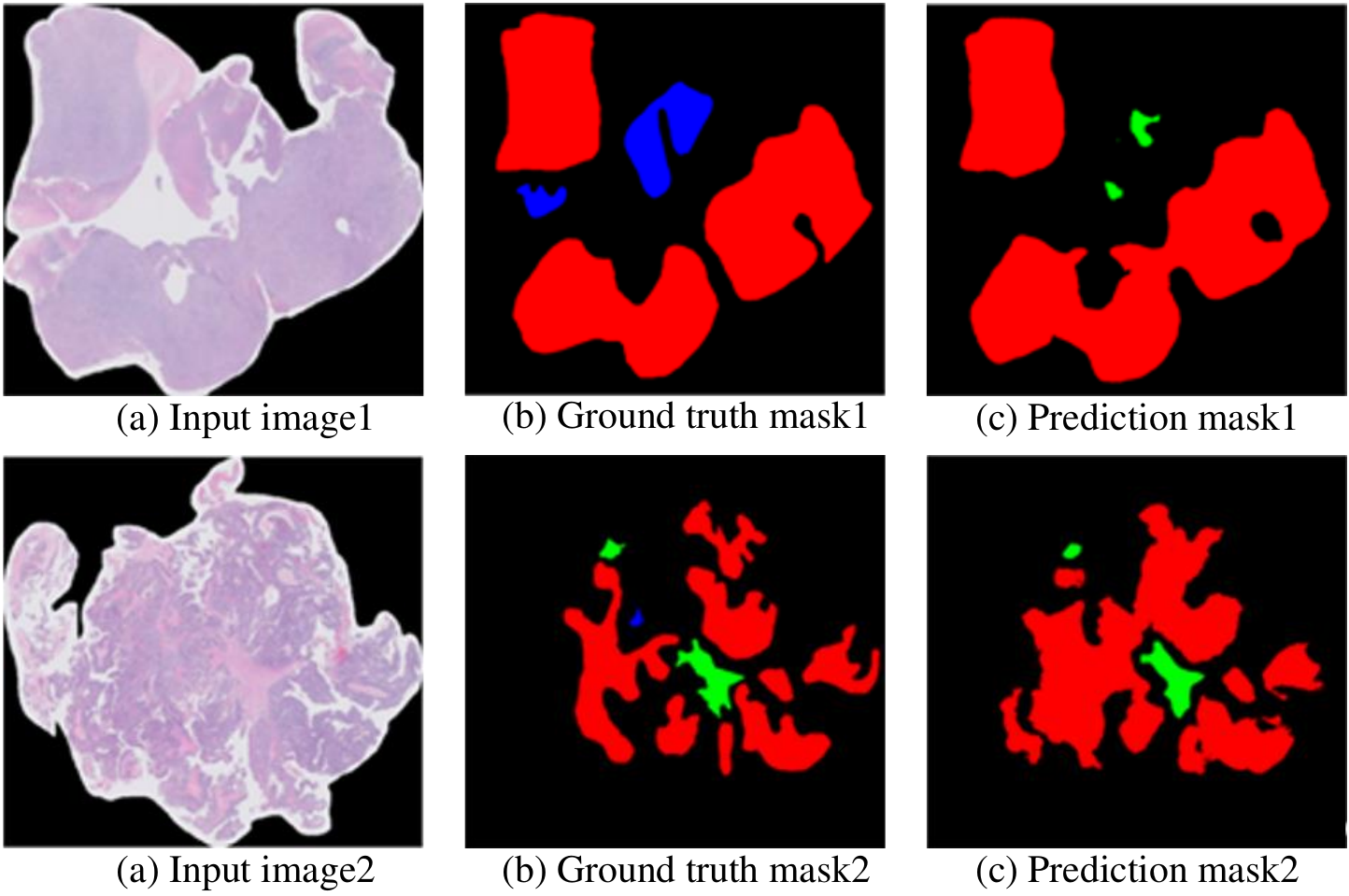}
      \caption{The reasoning effectiveness of segmentation}\label{segment}
    \end{figure}
    
\begin{table*}[t]
  \centering
  \begin{tabular}{@{}lccccccc@{}}
    \toprule
    \multirow{2}{*}{Strategy} & \multicolumn{2}{c}{UBC-OCEAN} & \multicolumn{2}{c}{BCNB} & \multicolumn{2}{c}{DigestPath2019} & Processing \\
    \cmidrule(lr){2-3} \cmidrule(lr){4-5} \cmidrule(lr){6-7}
    & AUC & Accuracy & AUC & Accuracy & AUC & Accuracy & Time(min) \\
    \midrule
    1. All patches & 0.945 & 0.929 & 0.953 & 0.941 & 0.959 & 0.942 & 57 \\
    2. Random patches & 0.931 & 0.920 & 0.937 & 0.929 & 0.942 & 0.937 & 21 \\
    3. Lesion area patches & \textbf{0.955} & \textbf{0.932} & \textbf{0.961} & \textbf{0.945} & \textbf{0.963} & \textbf{0.949} & 25 \\
    \bottomrule
  \end{tabular}
  \caption{Comparison of classification performance on different datasets using different patch selection strategies.}
  \label{tab:ablation}
\end{table*}
From our experiments, our segmentation network achieved an average IoU of 0.62 on the validation set, as can be seen from \cref{curves}(a), demonstrating effective and rapid screening of lesion areas at 1x low magnification. As shown in \cref{segment}, the red areas after segmentation are the lesion areas that need to be cropped for inference, and it can be seen that the area of patches requiring computation was reduced by approximately 50-70\%, greatly improving inference speed.
To verify the effect of pre-segmenting lesion areas on classification accuracy, we conducted ablation experiments: \textbf{1.} all non-background patches at multiple magnifications were input to our multi-scale fusion MIL network, \textbf{2.} randomly selected 3000 non-background multi-scale patches, with 2300, 570, and 130 patches at 20x, 10x, and 5x magnifications respectively, \textbf{3.} only selected lesion areas after segmentation for inference classification (average patch count after segmentation was 3000).
We selected a fixed random seed, using the same trained multi-scale MIL model and our feature extractor across different groups, calculating accuracy as the average of five-fold accuracies.

The experimental results in \cref{tab:ablation} show that when using lesion areas for multiple instance learning compared to using all patches, there was a slight improvement in classification accuracy (from 0.929 to 0.932 on UBC-OCEAN and from 0.941 to 0.945 on BCNB). However, compared to randomly selecting non-background patches, the lesion area segmentation strategy brought a significant accuracy improvement (from 0.920 to 0.932 on UBC-OCEAN and from 0.929 to 0.945 on BCNB).We conclude that for approaches using all patches, pre-segmenting lesion regions does not significantly improve classification accuracy. We hypothesize that our architecture can effectively handle all types of patches (including stroma, normal tissue, adipose tissue, and lesion regions), automatically filtering out interfering regions and achieving good feature fusion. However, in terms of time efficiency, our method significantly outperforms other approaches. Our calculated time includes the entire process from cropping WSI images to obtaining classification results through inference. Compared to the all\_patches method that directly processes all patches, we only need to crop and perform inference on the segmented lesion regions, greatly reducing computational load; compared to random cropping strategies, although our method adds segmentation and repositioning steps that take slightly longer, the overall efficiency remains superior. Specifically, our time savings come from two aspects: first, by only loading lesion regions, we reduce memory I/O overhead; second, by using only 50\% to 70\% of the original image area for feature extraction and inference, we significantly decrease computational requirements.

\subsection{Comparison with State-of-the-Art Methods}
To validate the effectiveness of our multi-scale multiple instance learning, we conducted comparative experiments against mainstream single-scale and multi-scale models. Our experiments performed binary classification on the BCNB and DigestPath2019 datasets, and five-class classification on the UBC-OCEAN dataset. For the BCNB and DigestPath2019 datasets, the MILbooster method utilized its proposed "bag filter." Notably, our approach incorporated the pre-segmentation module across all three datasets in our experiments.

\begin{table*}[t]
  \centering
  \begin{tabular}{@{}lcccccc@{}}
    \toprule
    \multirow{2}{*}{Method} & \multicolumn{2}{c}{UBC-OCEAN} & \multicolumn{2}{c}{BCNB} & \multicolumn{2}{c}{DigestPath2019} \\
    \cmidrule(lr){2-3} \cmidrule(lr){4-5} \cmidrule(lr){6-7}
    & AUC & Acc & AUC & Acc & AUC & Acc \\
    \midrule
    CSMIL$^{\ddagger}$ \cite{deng2022cross} & 0.929 & 0.899 & 0.922 & 0.897 & 0.928 & 0.913 \\
    MILbootser$^{\ddagger}$ \cite{qu2023boosting} & 0.944 & 0.922 & 0.947 & 0.902 & 0.946 & 0.924 \\
    DAS-MIL$^{\ddagger}$ \cite{bontempo2023graph} & 0.945 & 0.921 & 0.942 & 0.906 & 0.945 & 0.926 \\
    DSMIL$^{\ddagger}$ \cite{li2021dual} & 0.921 & 0.889 & 0.903 & 0.872 & 0.922 & 0.909 \\
    MambaMIL$^{\dagger}$ \cite{yang2024mambamil} & 0.913 & 0.879 & 0.899 & 0.874 & 0.907 & 0.878 \\
    TransMIL$^{\dagger}$ \cite{shao2021transmil} & 0.915 & 0.881 & 0.898 & 0.873 & 0.911 & 0.882 \\
    ABMIL$^{\dagger}$ \cite{ilse2018attention} & 0.898 & 0.862 & 0.884 & 0.851 & 0.896 & 0.859 \\
    \textbf{Ours}$^{\ddagger}$ & \textbf{0.955} & \textbf{0.932} & \textbf{0.961} & \textbf{0.945} & \textbf{0.963} & \textbf{0.949} \\
    \bottomrule
  \end{tabular}
  \caption{Comparison of classification performance of different MIL methods on UBC-OCEAN, BCNB, and DigestPath2019 datasets. $^{\dagger}$Single-scale methods. $^{\ddagger}$Multi-scale methods.}
  \label{tab:method_comparison}
\end{table*}

From the results in \cref{tab:method_comparison}, multi-scale multiple instance learning models generally outperform mainstream single-scale multiple instance learning models, demonstrating the importance of combining multi-scale information for pathological image analysis. When compared to other multi-scale methods, our approach still achieves state-of-the-art performance, proving the effectiveness of our multi-scale fusion MIL and the end-to-end trained feature extractor designed for it.

\subsection{Effectiveness Analysis of End-to-End Training Strategy}
       \begin{figure}[htbp]
      \centering
      \includegraphics[width=0.7\linewidth]{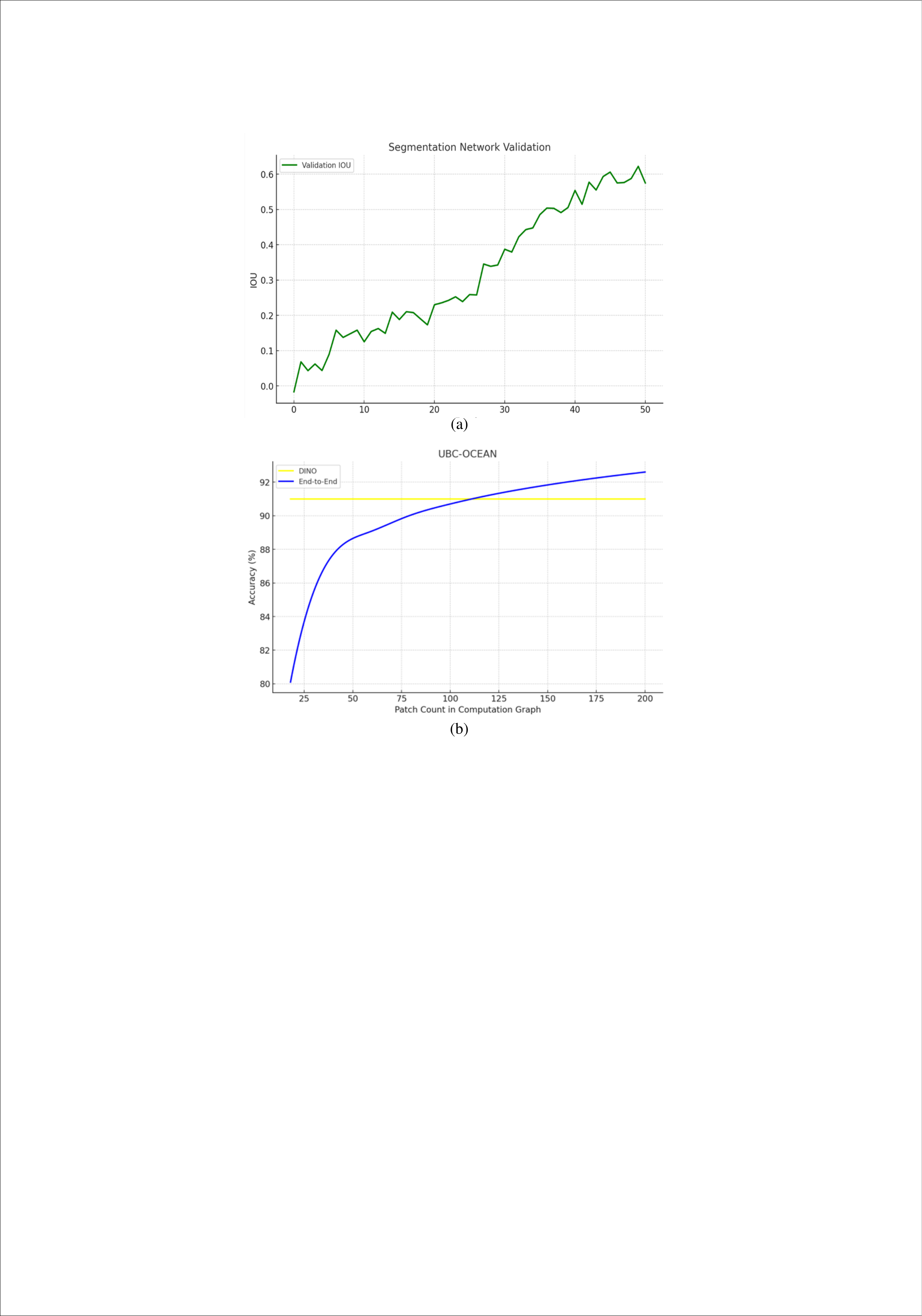}
      \caption{(a): Changes in IoU during the training process of U-Net++; (b): Comparison between self-supervised learning with DINO and end-to-end training methods.}\label{curves}
    \end{figure}
We compared the differences between self-supervised pretraining methods (DINO) and end-to-end training methods in multi-scale feature extraction and classification performance. Existing methods typically adopt a segmented training strategy, first pretraining the feature extractor through self-supervised methods, then using the MIL network for classification. This strategy leads to decoupling of optimization objectives between the feature extraction network and the MIL network, preventing collaborative optimization, resulting in insufficient adaptability between feature representation and classification tasks, and ultimately potentially causing the model to fall into suboptimal solutions. To verify the effectiveness of the end-to-end training strategy, we designed comparative experiments, using DINO contrastive learning to separately train feature extractors for three scales (20x, 10x, and 5x) for self-supervised pretraining, and compared with the end-to-end training strategy proposed in this paper. The MIL network structures for both methods were identical, and both underwent supplementary training of the MIL network in subsequent stages.
\begin{center}
  \begin{tabular}{@{}lccc@{}}
    \toprule
    \textbf{Methods} & \textbf{AUC} & \textbf{ACC} & \textbf{Time} \\
    \midrule
    Self-supervised & 0.950 & 0.928 & 71h \\
    End-to-end & 0.960 & 0.942 & 25h \\
    \bottomrule
  \end{tabular}
  \captionof{table}{Comparison cost time between self-supervised pretraining and end-to-end training approaches (averaged across datasets).}
  \label{tab:training_comparison}
\end{center}
As shown in Table~\ref{tab:training_comparison}, the end-to-end training method significantly outperforms the self-supervised pretraining method in both ACC and AUC metrics. Additionally, the training time of the end-to-end training method is significantly lower than that of the self-supervised pretraining method, further demonstrating its efficiency. We analyze that the end-to-end training method, by jointly optimizing the feature extractor and MIL network, achieves end-to-end alignment between the feature space and classification decision boundary, enabling the feature extractor to dynamically adapt to the aggregation needs of the MIL network, thereby enhancing the overall performance of the model. In contrast, the self-supervised pretraining method, due to the decoupling of feature extraction and classification tasks, has insufficient adaptability between feature representation and classification tasks, and heterogeneous feature distributions across scales, resulting in limited performance.

\subsection{Effect of Batch Size in End-to-End Training}
In end-to-end training, when the sequence includes a larger number of patches, the model can fully utilize multi-scale contextual information, obtaining more critical discriminative features from the WSI image, thereby enhancing the capability of feature extraction and classification performance. However, when the sequence contains only one patch, this method degrades to a single-scale instance-based approach \cite{campanella2019clinical,chikontwe2020multiple,kanavati2020weakly,xu2019camel}, where each instance is directly assigned a WSI-level pseudo-label. Under this setting, the model's learning process may introduce more noise, making it difficult for the feature extractor to effectively learn discriminative features. Therefore, we set up experiments to vary the number of patches included in the computational graph during end-to-end training of the feature extractor, comparing with the method of using DINO self-supervised training for three scales, comparing ACC on the UBC-OCEAN dataset. Note that after the feature extractors were trained, both methods additionally trained the MIL network in the context of long sequences.

According to the curve in \Cref{curves}(b), we conclude that using the end-to-end training method, when there are few instances in the MIL input sequence, the feature extractor will learn a large amount of data noise, showing poor performance. When approximately 120 images are retained in the computational graph, the performance of the end-to-end training method is about equal to that of the DINO self-supervised method.

\section{Conclusion}

This paper presents MsaMIL-Net, an end-to-end multi-scale multi-instance learning network framework for efficient WSI classification. By integrating multi-scale feature extraction with multi-instance learning, the framework significantly enhances WSI classification performance, achieving state-of-the-art accuracy and AUC metrics across multiple public datasets. Unlike existing segmented training methods, MsaMIL-Net implements an end-to-end training strategy that enables collaborative optimization between the feature extraction network and MIL network, ensuring maximum alignment between feature representation and classification tasks.

Additionally, the proposed Semantic Feature Filtering Module (SFFM) and Multi-Scale Feature Extraction Module (MSFEM) effectively reduce interference from non-lesion regions, improve computational efficiency, while achieving global modeling and efficient fusion of multi-scale features through the multi-scale fusion MIL module. Experimental results demonstrate that MsaMIL-Net performs excellently on three datasets: DigestPath2019, BCNB, and UBC-OCEAN, significantly outperforming existing single-scale and multi-scale MIL methods in classification accuracy and AUC metrics.

Specifically, MsaMIL-Net achieves an AUC of 0.955 on the UBC-OCEAN dataset, approximately 2.6\% higher than CSMIL (0.929) and 1\% higher than DAS-MIL (0.945); on the BCNB dataset, it reaches an AUC of 0.961, about 3.9\% higher than CSMIL (0.922) and 1.9\% higher than DAS-MIL (0.942). Furthermore, the end-to-end training strategy outperforms traditional self-supervised pre-training methods in both training time and classification performance, further validating its efficiency and superiority.

Through ablation and comparative experiments, we verify the effectiveness of multi-scale feature extraction and end-to-end training strategies, demonstrating MsaMIL-Net's powerful capability in handling complex WSI classification tasks. Future research directions may include further optimization of multi-scale feature extraction and fusion mechanisms, exploration of more efficient feature filtering methods, and extension of MsaMIL-Net to other medical image analysis tasks.
{
    \small
    \bibliographystyle{ieeenat_fullname}
    \bibliography{main}

\begin{thebibliography}{31}
\providecommand{\natexlab}[1]{#1}
\providecommand{\url}[1]{\texttt{#1}}
\expandafter\ifx\csname urlstyle\endcsname\relax
  \providecommand{\doi}[1]{doi: #1}\else
  \providecommand{\doi}{doi: \begingroup \urlstyle{rm}\Url}\fi

\bibitem[Asadi-Aghbolaghi et~al.(2024)Asadi-Aghbolaghi, Farahani, Zhang, Akbari, Kim, Chow, Dane, Consortium, Consortium, Huntsman, et~al.]{asadi2024machine}
Maryam Asadi-Aghbolaghi, Hossein Farahani, Allen Zhang, Ardalan Akbari, Sirim Kim, Ashley Chow, Sohier Dane, OCEAN~Challenge Consortium, OTTA Consortium, David~G Huntsman, et~al.
\newblock Machine learning-driven histotype diagnosis of ovarian carcinoma: insights from the ocean ai challenge.
\newblock \emph{medRxiv}, pages 2024--04, 2024.

\bibitem[Bejnordi et~al.(2017)Bejnordi, Veta, Van~Diest, Van~Ginneken, Karssemeijer, Litjens, Van Der~Laak, Hermsen, Manson, Balkenhol, et~al.]{bejnordi2017diagnostic}
Babak~Ehteshami Bejnordi, Mitko Veta, Paul~Johannes Van~Diest, Bram Van~Ginneken, Nico Karssemeijer, Geert Litjens, Jeroen~AWM Van Der~Laak, Meyke Hermsen, Quirine~F Manson, Maschenka Balkenhol, et~al.
\newblock Diagnostic assessment of deep learning algorithms for detection of lymph node metastases in women with breast cancer.
\newblock \emph{Jama}, 318\penalty0 (22):\penalty0 2199--2210, 2017.

\bibitem[Bontempo et~al.(2023)Bontempo, Bolelli, Porrello, Calderara, and Ficarra]{bontempo2023graph}
Gianpaolo Bontempo, Federico Bolelli, Angelo Porrello, Simone Calderara, and Elisa Ficarra.
\newblock A graph-based multi-scale approach with knowledge distillation for wsi classification.
\newblock \emph{IEEE Transactions on Medical Imaging}, 43\penalty0 (4):\penalty0 1412--1421, 2023.

\bibitem[Campanella et~al.(2019)Campanella, Hanna, Geneslaw, Miraflor, Werneck Krauss~Silva, Busam, Brogi, Reuter, Klimstra, and Fuchs]{campanella2019clinical}
Gabriele Campanella, Matthew~G Hanna, Luke Geneslaw, Allen Miraflor, Vitor Werneck Krauss~Silva, Klaus~J Busam, Edi Brogi, Victor~E Reuter, David~S Klimstra, and Thomas~J Fuchs.
\newblock Clinical-grade computational pathology using weakly supervised deep learning on whole slide images.
\newblock \emph{Nature medicine}, 25\penalty0 (8):\penalty0 1301--1309, 2019.

\bibitem[Chikontwe et~al.(2020)Chikontwe, Kim, Nam, Go, and Park]{chikontwe2020multiple}
Philip Chikontwe, Meejeong Kim, Soo~Jeong Nam, Heounjeong Go, and Sang~Hyun Park.
\newblock Multiple instance learning with center embeddings for histopathology classification.
\newblock In \emph{Medical Image Computing and Computer Assisted Intervention--MICCAI 2020: 23rd International Conference, Lima, Peru, October 4--8, 2020, Proceedings, Part V 23}, pages 519--528. Springer, 2020.

\bibitem[Da et~al.(2022)Da, Huang, Li, Zuo, Zhang, Liu, Chen, Li, Xu, Hu, et~al.]{da2022digestpath}
Qian Da, Xiaodi Huang, Zhongyu Li, Yanfei Zuo, Chenbin Zhang, Jingxin Liu, Wen Chen, Jiahui Li, Dou Xu, Zhiqiang Hu, et~al.
\newblock Digestpath: A benchmark dataset with challenge review for the pathological detection and segmentation of digestive-system.
\newblock \emph{Medical image analysis}, 80:\penalty0 102485, 2022.

\bibitem[Deng et~al.(2022)Deng, Cui, Remedios, Bao, Womick, Chiron, Li, Roland, Lau, Liu, et~al.]{deng2022cross}
Ruining Deng, Can Cui, Lucas~W Remedios, Shunxing Bao, R~Michael Womick, Sophie Chiron, Jia Li, Joseph~T Roland, Ken~S Lau, Qi Liu, et~al.
\newblock Cross-scale attention guided multi-instance learning for crohn’s disease diagnosis with pathological images.
\newblock In \emph{International workshop on multiscale multimodal medical imaging}, pages 24--33. Springer, 2022.

\bibitem[Fang et~al.(2024)Fang, Wang, Zhang, Wang, Zhang, Ji, and Zhang]{fang2024mammil}
Zijie Fang, Yifeng Wang, Ye Zhang, Zhi Wang, Jian Zhang, Xiangyang Ji, and Yongbing Zhang.
\newblock Mammil: Multiple instance learning for whole slide images with state space models.
\newblock In \emph{2024 IEEE International Conference on Bioinformatics and Biomedicine (BIBM)}, pages 3200--3205. IEEE, 2024.

\bibitem[Farahani et~al.(2022)Farahani, Boschman, Farnell, Darbandsari, Zhang, Ahmadvand, Jones, Huntsman, K{\"o}bel, Gilks, et~al.]{farahani2022deep}
Hossein Farahani, Jeffrey Boschman, David Farnell, Amirali Darbandsari, Allen Zhang, Pouya Ahmadvand, Steven~JM Jones, David Huntsman, Martin K{\"o}bel, C~Blake Gilks, et~al.
\newblock Deep learning-based histotype diagnosis of ovarian carcinoma whole-slide pathology images.
\newblock \emph{Modern Pathology}, 35\penalty0 (12):\penalty0 1983--1990, 2022.

\bibitem[Filiot et~al.(2024)Filiot, Jacob, Mac~Kain, and Saillard]{filiot2024phikon}
Alexandre Filiot, Paul Jacob, Alice Mac~Kain, and Charlie Saillard.
\newblock Phikon-v2, a large and public feature extractor for biomarker prediction.
\newblock \emph{arXiv preprint arXiv:2409.09173}, 2024.

\bibitem[Hashimoto et~al.(2020)Hashimoto, Fukushima, Koga, Takagi, Ko, Kohno, Nakaguro, Nakamura, Hontani, and Takeuchi]{hashimoto2020multi}
Noriaki Hashimoto, Daisuke Fukushima, Ryoichi Koga, Yusuke Takagi, Kaho Ko, Kei Kohno, Masato Nakaguro, Shigeo Nakamura, Hidekata Hontani, and Ichiro Takeuchi.
\newblock Multi-scale domain-adversarial multiple-instance cnn for cancer subtype classification with unannotated histopathological images.
\newblock In \emph{Proceedings of the IEEE/CVF conference on computer vision and pattern recognition}, pages 3852--3861, 2020.

\bibitem[Ilse et~al.(2018)Ilse, Tomczak, and Welling]{ilse2018attention}
Maximilian Ilse, Jakub Tomczak, and Max Welling.
\newblock Attention-based deep multiple instance learning.
\newblock In \emph{International conference on machine learning}, pages 2127--2136. PMLR, 2018.

\bibitem[Kanavati et~al.(2020)Kanavati, Toyokawa, Momosaki, Rambeau, Kozuma, Shoji, Yamazaki, Takeo, Iizuka, and Tsuneki]{kanavati2020weakly}
Fahdi Kanavati, Gouji Toyokawa, Seiya Momosaki, Michael Rambeau, Yuka Kozuma, Fumihiro Shoji, Koji Yamazaki, Sadanori Takeo, Osamu Iizuka, and Masayuki Tsuneki.
\newblock Weakly-supervised learning for lung carcinoma classification using deep learning.
\newblock \emph{Scientific reports}, 10\penalty0 (1):\penalty0 9297, 2020.

\bibitem[Li et~al.(2021)Li, Li, and Eliceiri]{li2021dual}
Bin Li, Yin Li, and Kevin~W Eliceiri.
\newblock Dual-stream multiple instance learning network for whole slide image classification with self-supervised contrastive learning.
\newblock In \emph{Proceedings of the IEEE/CVF conference on computer vision and pattern recognition}, pages 14318--14328, 2021.

\bibitem[Lu et~al.(2021)Lu, Chen, Williamson, Zhao, Shady, Lipkova, and Mahmood]{lu2021ai}
Ming~Y Lu, Tiffany~Y Chen, Drew~FK Williamson, Melissa Zhao, Maha Shady, Jana Lipkova, and Faisal Mahmood.
\newblock Ai-based pathology predicts origins for cancers of unknown primary.
\newblock \emph{Nature}, 594\penalty0 (7861):\penalty0 106--110, 2021.

\bibitem[Lu et~al.(2022)Lu, Chen, Kong, Lipkova, Singh, Williamson, Chen, and Mahmood]{lu2022federated}
Ming~Y Lu, Richard~J Chen, Dehan Kong, Jana Lipkova, Rajendra Singh, Drew~FK Williamson, Tiffany~Y Chen, and Faisal Mahmood.
\newblock Federated learning for computational pathology on gigapixel whole slide images.
\newblock \emph{Medical image analysis}, 76:\penalty0 102298, 2022.

\bibitem[Mahmood et~al.(2019)Mahmood, Borders, Chen, McKay, Salimian, Baras, and Durr]{mahmood2019deep}
Faisal Mahmood, Daniel Borders, Richard~J Chen, Gregory~N McKay, Kevan~J Salimian, Alexander Baras, and Nicholas~J Durr.
\newblock Deep adversarial training for multi-organ nuclei segmentation in histopathology images.
\newblock \emph{IEEE transactions on medical imaging}, 39\penalty0 (11):\penalty0 3257--3267, 2019.

\bibitem[Oquab et~al.(2023)Oquab, Darcet, Moutakanni, Vo, Szafraniec, Khalidov, Fernandez, Haziza, Massa, El-Nouby, et~al.]{oquab2023dinov2}
Maxime Oquab, Timoth{\'e}e Darcet, Th{\'e}o Moutakanni, Huy Vo, Marc Szafraniec, Vasil Khalidov, Pierre Fernandez, Daniel Haziza, Francisco Massa, Alaaeldin El-Nouby, et~al.
\newblock Dinov2: Learning robust visual features without supervision.
\newblock \emph{arXiv preprint arXiv:2304.07193}, 2023.

\bibitem[Qu et~al.(2022{\natexlab{a}})Qu, Liu, Liu, Wang, and Song]{qu2022towards}
Linhao Qu, Siyu Liu, Xiaoyu Liu, Manning Wang, and Zhijian Song.
\newblock Towards label-efficient automatic diagnosis and analysis: a comprehensive survey of advanced deep learning-based weakly-supervised, semi-supervised and self-supervised techniques in histopathological image analysis.
\newblock \emph{Physics in Medicine \& Biology}, 67\penalty0 (20):\penalty0 20TR01, 2022{\natexlab{a}}.

\bibitem[Qu et~al.(2022{\natexlab{b}})Qu, Luo, Liu, Wang, and Song]{qu2022dgmil}
Linhao Qu, Xiaoyuan Luo, Shaolei Liu, Manning Wang, and Zhijian Song.
\newblock Dgmil: Distribution guided multiple instance learning for whole slide image classification.
\newblock In \emph{International conference on medical image computing and computer-assisted intervention}, pages 24--34. Springer, 2022{\natexlab{b}}.

\bibitem[Qu et~al.(2022{\natexlab{c}})Qu, Wang, Song, et~al.]{qu2022bi}
Linhao Qu, Manning Wang, Zhijian Song, et~al.
\newblock Bi-directional weakly supervised knowledge distillation for whole slide image classification.
\newblock \emph{Advances in Neural Information Processing Systems}, 35:\penalty0 15368--15381, 2022{\natexlab{c}}.

\bibitem[Qu et~al.(2023)Qu, Yang, Duan, Ma, Wang, Wang, and Song]{qu2023boosting}
Linhao Qu, Zhiwei Yang, Minghong Duan, Yingfan Ma, Shuo Wang, Manning Wang, and Zhijian Song.
\newblock Boosting whole slide image classification from the perspectives of distribution, correlation and magnification.
\newblock In \emph{Proceedings of the IEEE/CVF International Conference on Computer Vision}, pages 21463--21473, 2023.

\bibitem[Qu et~al.(2024)Qu, Ma, Luo, Guo, Wang, and Song]{qu2024rethinking}
Linhao Qu, Yingfan Ma, Xiaoyuan Luo, Qinhao Guo, Manning Wang, and Zhijian Song.
\newblock Rethinking multiple instance learning for whole slide image classification: A good instance classifier is all you need.
\newblock \emph{IEEE Transactions on Circuits and Systems for Video Technology}, 2024.

\bibitem[Shao et~al.(2021)Shao, Bian, Chen, Wang, Zhang, Ji, et~al.]{shao2021transmil}
Zhuchen Shao, Hao Bian, Yang Chen, Yifeng Wang, Jian Zhang, Xiangyang Ji, et~al.
\newblock Transmil: Transformer based correlated multiple instance learning for whole slide image classification.
\newblock \emph{Advances in neural information processing systems}, 34:\penalty0 2136--2147, 2021.

\bibitem[Srinidhi et~al.(2021)Srinidhi, Ciga, and Martel]{srinidhi2021deep}
Chetan~L Srinidhi, Ozan Ciga, and Anne~L Martel.
\newblock Deep neural network models for computational histopathology: A survey.
\newblock \emph{Medical image analysis}, 67:\penalty0 101813, 2021.

\bibitem[Vaswani et~al.(2017)Vaswani, Shazeer, Parmar, Uszkoreit, Jones, Gomez, Kaiser, and Polosukhin]{vaswani2017attention}
Ashish Vaswani, Noam Shazeer, Niki Parmar, Jakob Uszkoreit, Llion Jones, Aidan~N Gomez, {\L}ukasz Kaiser, and Illia Polosukhin.
\newblock Attention is all you need.
\newblock \emph{Advances in neural information processing systems}, 30, 2017.

\bibitem[Wang et~al.(2022)Wang, Yang, Zhang, Wang, Zhang, Yang, Huang, and Han]{wang2022transformer}
Xiyue Wang, Sen Yang, Jun Zhang, Minghui Wang, Jing Zhang, Wei Yang, Junzhou Huang, and Xiao Han.
\newblock Transformer-based unsupervised contrastive learning for histopathological image classification.
\newblock \emph{Medical image analysis}, 81:\penalty0 102559, 2022.

\bibitem[Xu et~al.(2019)Xu, Song, Sun, Ku, Yang, Liu, Wang, Ma, and Xu]{xu2019camel}
Gang Xu, Zhigang Song, Zhuo Sun, Calvin Ku, Zhe Yang, Cancheng Liu, Shuhao Wang, Jianpeng Ma, and Wei Xu.
\newblock Camel: A weakly supervised learning framework for histopathology image segmentation.
\newblock In \emph{Proceedings of the IEEE/CVF International Conference on computer vision}, pages 10682--10691, 2019.

\bibitem[Yang et~al.(2024)Yang, Wang, and Chen]{yang2024mambamil}
Shu Yang, Yihui Wang, and Hao Chen.
\newblock Mambamil: Enhancing long sequence modeling with sequence reordering in computational pathology.
\newblock In \emph{International Conference on Medical Image Computing and Computer-Assisted Intervention}, pages 296--306. Springer, 2024.

\bibitem[Zhang et~al.(2022)Zhang, Wang, Li, and Liu]{zhang2022self}
Haohuan Zhang, Ruixuan Wang, Chunxiao Li, and Hongmei Liu.
\newblock Self-supervised learning and multi-scale ensemble for ebv prediction based on whole slide image of gastric cancer.
\newblock In \emph{Third International Conference on Computer Science and Communication Technology (ICCSCT 2022)}, pages 903--907. SPIE, 2022.

\bibitem[Zhou et~al.(2018)Zhou, Rahman~Siddiquee, Tajbakhsh, and Liang]{zhou2018unet++}
Zongwei Zhou, Md~Mahfuzur Rahman~Siddiquee, Nima Tajbakhsh, and Jianming Liang.
\newblock Unet++: A nested u-net architecture for medical image segmentation.
\newblock In \emph{Deep learning in medical image analysis and multimodal learning for clinical decision support: 4th international workshop, DLMIA 2018, and 8th international workshop, ML-CDS 2018, held in conjunction with MICCAI 2018, Granada, Spain, September 20, 2018, proceedings 4}, pages 3--11. Springer, 2018.

\end{thebibliography}
}

\end{document}